\documentclass[nonacm, sigconf]{acmart}

\setcopyright{none}
\makeatletter
\renewcommand\@formatdoi[1]{\ignorespaces}
\makeatother
\renewcommand\footnotetextcopyrightpermission[1]{}

\usepackage{times}
\usepackage{url}
\usepackage{latexsym}

\usepackage{amsthm}
\usepackage{mathtools}
\usepackage{graphicx}
\usepackage{algorithm}
\usepackage{algorithmic}
\usepackage{multirow}
\usepackage{wrapfig}

\AtBeginDocument{%
  \providecommand\BibTeX{{%
    \normalfont B\kern-0.5em{\scshape i\kern-0.25em b}\kern-0.8em\TeX}}}

\begin{document}

\title{Entity Context Graph: Learning Entity Representations from Semi-Structured Textual Sources on the Web}


\author{Kalpa Gunaratna}
\affiliation{%
  \institution{Samsung Research America}
}
\email{k.gunaratna@samsung.com}

\author{Yu Wang}
\affiliation{%
  \institution{Samsung Research America}
}
\email{yu.wang1@samsung.com}

\author{Hongxia Jin}
\affiliation{%
  \institution{Samsung Research America}
}
\email{hongxia.jin@samsung.com}








\begin{abstract}
Knowledge is captured in the form of entities and their relationships and stored in knowledge graphs. Knowledge graphs enhance the capabilities of applications in many different areas including Web search, recommendation, and natural language understanding. This is mainly because, entities enable machines to understand things that go beyond simple tokens. Many modern algorithms use learned entity embeddings from these structured representations. However, building a knowledge graph takes time and effort, hence very costly and nontrivial. On the other hand, many Web sources describe entities in some structured format and therefore, finding ways to get them into useful entity knowledge is advantageous. We propose an approach that processes entity centric textual knowledge sources to learn entity embeddings and in turn avoids the need for a traditional knowledge graph. We first extract triples into the new representation format that does not use traditional complex triple extraction methods defined by pre-determined relationship labels. Then we learn entity embeddings through this new type of triples. We show that the embeddings learned from our approach are: (i) high quality and comparable to a known knowledge graph-based embeddings and can be used to improve them further, (ii) better than a contextual language model-based entity embeddings, and (iii) easy to compute and versatile in domain-specific applications where a knowledge graph is not readily available.

\end{abstract}



\maketitle

\section{Introduction}
Knowledge stored in machine process-able formats help machines to have great level of intelligence. Knowledge Graph (KG)~\footnote{KG refers to a well-formed structured knowledge graph in this paper - e.g., Freebase and DBpedia.} is a way of encoding human and world knowledge in an easy-to-process structured format. They typically use knowledge representation languages like Resource Description Framework (RDF)~\cite{manola2004rdf}. A KG primarily consists of entities (i.e., things that can be described) and relationships that link pairs of entities to form triples. A triple (head entity $h$, relationship $r$, tail entity $t$) is the basic building block of this multi-relational knowledge model. Lately, understanding the importance of knowledge led to the  creation of many encyclopedic KGs (e.g., Yago~\cite{suchanek2007yago}, DBpedia~\cite{auer2007dbpedia}, and Freebase~\cite{bollacker2008freebase}) and domain-specific KGs (e.g., see many available in Linked Open Data cloud~\footnote{https://lod-cloud.net/}). While symbolic knowledge in KGs can be directly used in some question answering, reasoning, Web search and retrieval techniques, modern deep learning-based algorithms require the entity inputs to be in numeric vector format. Representation Leaning (LR) methods (e.g., ~\cite{bordes2013translating,bordes2014semantic,nickel2011three,yang2014embedding,ristoski2016rdf2vec,nickel2016holographic,wang2017knowledge}) have been introduced to convert the `symbolic' knowledge in KGs  into `numeric' embedding vectors. In recent developments, it has been shown that deep learning applications can be further improved by incorporating knowledge (e.g., improving contextual language models using entity embeddings learned from a KG~\cite{zhang2019ernie}). However, these approaches assume that high quality KGs are readily available to learn the entity embeddings.

Building a KG is resource-heavy and time-consuming and hence, costly. For example, the cost of building the popular Freebase - KG is estimated to be 6.75 billion U.S. dollars~\cite{paulheimmuch}. Some encyclopedic KGs like DBpedia and Yago are extracted from high quality semi-structured data on the Web (e.g., infoboxes in Wikipedia and WordNet~\cite{miller1995wordnet} descriptions) to reduce manual labor in verifying the facts that go into the KGs. This type of KG creation depends on very specific and verified data. On the other hand, approaches that can extract triples from text documents to construct KGs still require high quality manually input seed patterns and schema/ontology based rules (e.g., NELL~\cite{carlson2010toward} and DeepDive~\cite{niu2012deepdive}). Therefore, RL techniques are not as versatile as one might think in real applications as they heavily depend on the availability of high quality KGs. However, there exists textual knowledge sources (e.g., Wikipedia, product catalogues, reviews, etc.) that focus on describing entities in detail. Our approach uses these sources instead of high quality structured KGs to learn entity embeddings.

\begin{figure}
    \centering
    \includegraphics[scale=0.45, trim=0.2cm 11.5cm 4.8cm 0cm, clip=true]{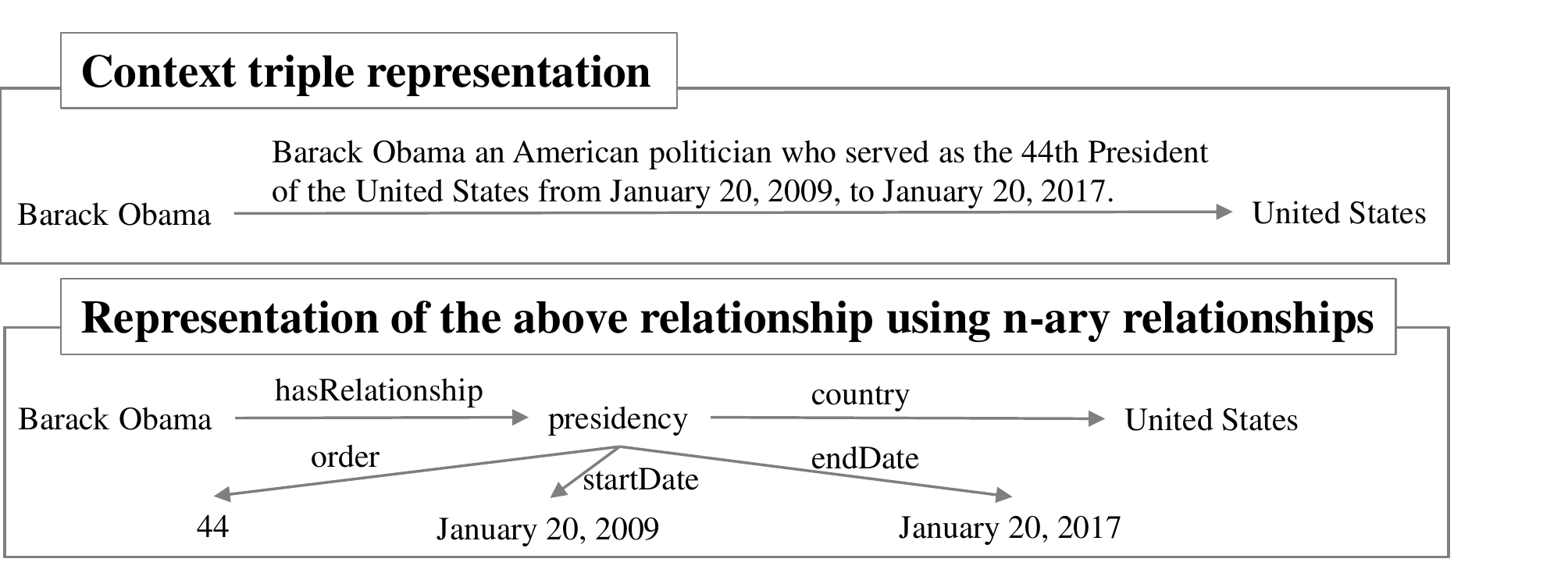}
    \caption{Comparison of ECG and KG triples. ECGs have less complex structure and easy to extract while KG triples have to use complex triple patterns to represent the same knowledge.}
    \label{fig_ecg_triple}
\end{figure}

Representation learning methods process KG triples that consist of a small fixed set of relations. In contrast, in our proposed approach, we allow relations between entity pairs to be textual descriptions (see Figure~\ref{fig_ecg_triple}) that form `context triples'. This new and relaxed triple representation format creates an entity graph called `Entity Context Graph' (ECG). In fact, a context triple can easily represent a complex data/ontology pattern because of its contextual relationship description whereas a KG-type representation may have to follow a complex data/ontology pattern to represent the same information. For example, the context triple shown in Figure~\ref{fig_ecg_triple} requires an n-ary relationship in RDF. Because of its simplicity, building an ECG is easy, efficient, and completely automated compared to constructing a KG. Further, to the best of our knowledge, using our easy to represent context triple format to learn entity embeddings is different to existing approaches. Our contributions in this paper are listed as follows:\\
\begin{enumerate}
    \item We design a system which can automatically build entity context graphs from semi-structured entity-centric knowledge sources and compute entity embeddings without needing to process a KG.
    \item We show that the results generated by ECG-based entity embeddings are comparable to purely KG-based ones, and better than a contextual language model-based entity embedding approach.
    \item In addition, we demonstrate that ECGs can improve KG entity embedding quality by adapting our method to jointly learn from both KG triples and context triples without any alignment to the KG triples.
\end{enumerate}

We conducted comprehensive evaluations and analysis to show that our approach is capable of computing quality embeddings and versatile in domain-specific use cases where finding a relevant KG is impossible. The rest of the paper is organized as follows. Section~\ref{related_work} presents related work. Section~\ref{approach} describes our approach and Section~\ref{evaluation} presents the evaluation results. Section~\ref{general_discussion} provides a general discussion with future directions and we conclude with final remarks in Section~\ref{conclusion}.

\section{Related Work}
\label{related_work}

The structured and symbolic format of KGs can be transformed into embedding space using RL algorithms (e.g., ~\cite{bordes2013translating,ristoski2016rdf2vec,toutanova2015representing,neelakantan2015compositional,wang2014knowledge,lin2015learning,ji2015knowledge,ji2016knowledge,ijcai2017-183,wang2017knowledge}). Bordes et al.~\cite{bordes2013translating} proposed TransE that translates entities using relations in the embedding space. The approach makes use of individual triples in the graphs to learn embeddings for both relationships and entities. Following TransE, many variants were proposed to improve its capabilities like TransH~\cite{wang2014knowledge}, TransR~\cite{lin2015learning}, TransD~\cite{ji2015knowledge}, TransSparse~\cite{ji2016knowledge}, STransE~\cite{nguyen2016stranse}, etc. Some of the other recent approaches include matrix factorization~\cite{trouillon2016complex}, graph convolutions~\cite{schlichtkrull2018modeling}, convolutional networks~\cite{dettmers2018convolutional}, and cross-over links~\cite{zhang2019interaction}.

Embedding learning on KGs does not work without a KG. There are several ways to build a KG: manual and community based (e.g., Freebase~\cite{bollacker2008freebase} and Wikidata~\cite{vrandevcic2014wikidata}); extracted from well defined semi-structured knowledge sources (e.g., DBpedia~\cite{auer2007dbpedia} and Yago~\cite{suchanek2007yago} extracted using Wikipedia infoboxes and WordNets); or learned to extract triples using supervised and incremental machine learning and domain expert input seed patterns(e.g., NELL~\cite{carlson2010toward} and DeepDive~\cite{shin2015incremental}). In all cases, they always consist of seed ontology patterns (mainly to detect relationships) and undergo thorough manual verification to ensure quality. This makes KGs costly~\cite{paulheimmuch} and hard to build. For example, systems like NELL and DeepDive that are proposed to extract KG triples from text documents still require seed patterns/ontology rules to guide them. On the other hand, there exists contextual word embedding approaches (e.g., BERT~\cite{devlin2018bert}) and KGs are used to improve their performances (e.g., ERNIE~\cite{zhang2019ernie} and K-BERT~\cite{liu2020k}). ERNIE uses entity masking and K-BERT processes full KG triples to improve the contextual language model. Even though transformer-based language models are achieving superior results in predicting/encoding words, their use in pure entity embedding learning has not been studied well without an already existing KG. We show that our method captures entity embeddings better than ERNIE (adapted to learn entity embeddings).

Different to all the above, our approach extracts different type of triples automatically to form ECGs from entity-centric knowledge sources. These triples have text descriptions as relationships compared to the labels (i.e., short phrases) in KG triples. Use of many surface forms of relationships ~\cite{riedel2013relation} to model relationships between two entities and CNN-based sentence attention for relation extraction ~\cite{lin2016neural} are some early works related to inferring relationships for KG completion and creation. These types of work are also orthogonal to our approach as we diverge from label-based relationship representation. To the best of our knowledge, this is the first such approach proposed to learn entity embeddings using triples that contain lengthy free-form text as relationships. We learn entity embeddings by adapting TransE~\cite{bordes2013translating} on the extracted triples and encoding the lengthy relationship texts, that then works similar to learning embeddings on a KG. Our method is useful in domain specific applications where KGs do not exist. Further, our extraction of ECGs can complement KGs in embedding learning compared to text + KG embedding learning approaches~\cite{wang2016text,an2018accurate} without any alignment or attention mechanism.

\section{Proposed Approach}
\label{approach}

\begin{figure}
    \centering
    \includegraphics[scale=0.7, trim=0cm 5.9cm 14cm 0.1cm, clip=true]{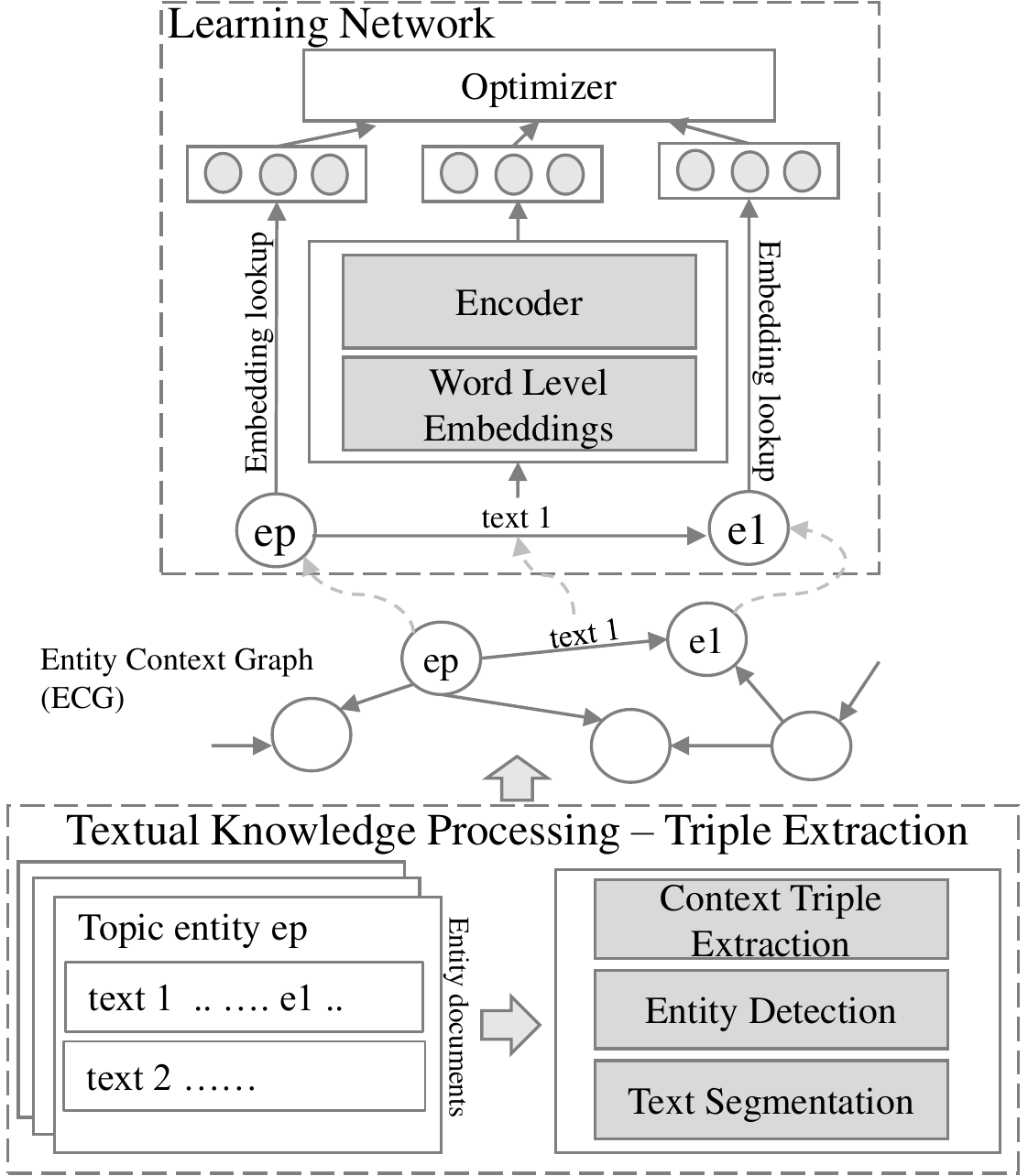}
    \caption{Proposed approach architecture. First, context triples are extracted from an entity based source. Then the context triples are used in the embedding learning network with a relation encoder to translate the relation text into vector space.}
    \label{fig_framework}
\end{figure}

In this work, we propose a method to compute entity embeddings using `entity-relation-entity' representation model, which makes graphs. Graph is a natural way of representing links between entities to capture semantics of the entities. Knowledge graph is a graph where the links are clearly defined (i.e., labeled). Similar to traditional knowledge representation, we pay close attention to relationships that link entities with meaning~\cite{sheth2004relationships} because, relationships are the key to capture semantics of the entities. We first extract a graph representation from a given textual entity centric dataset and operate on this graph to compute the embeddings. Our graph links are also labeled in the sense that it contains longer text compared to short phrases in a traditional KG. Therefore, we avoid the costly process of building a state-of-the-art KG but retain the ability to model and learn multi-relational nature of the entities~\footnote{Various relationship types among entities define the semantics of the entities in a KG (i.e., semantic graph).} in embedding learning. The proposed approach (see Figure~\ref{fig_framework}) consists of two main components: (i) context triple extraction to build ECGs and (ii) embedding learning for entities using ECGs. Next, we describe them in detail.

\subsection{Triple Extraction}

\begin{figure}
    \centering
    \includegraphics[scale=0.33, trim=0.1cm 5cm 0.1cm 0.1cm, clip=true]{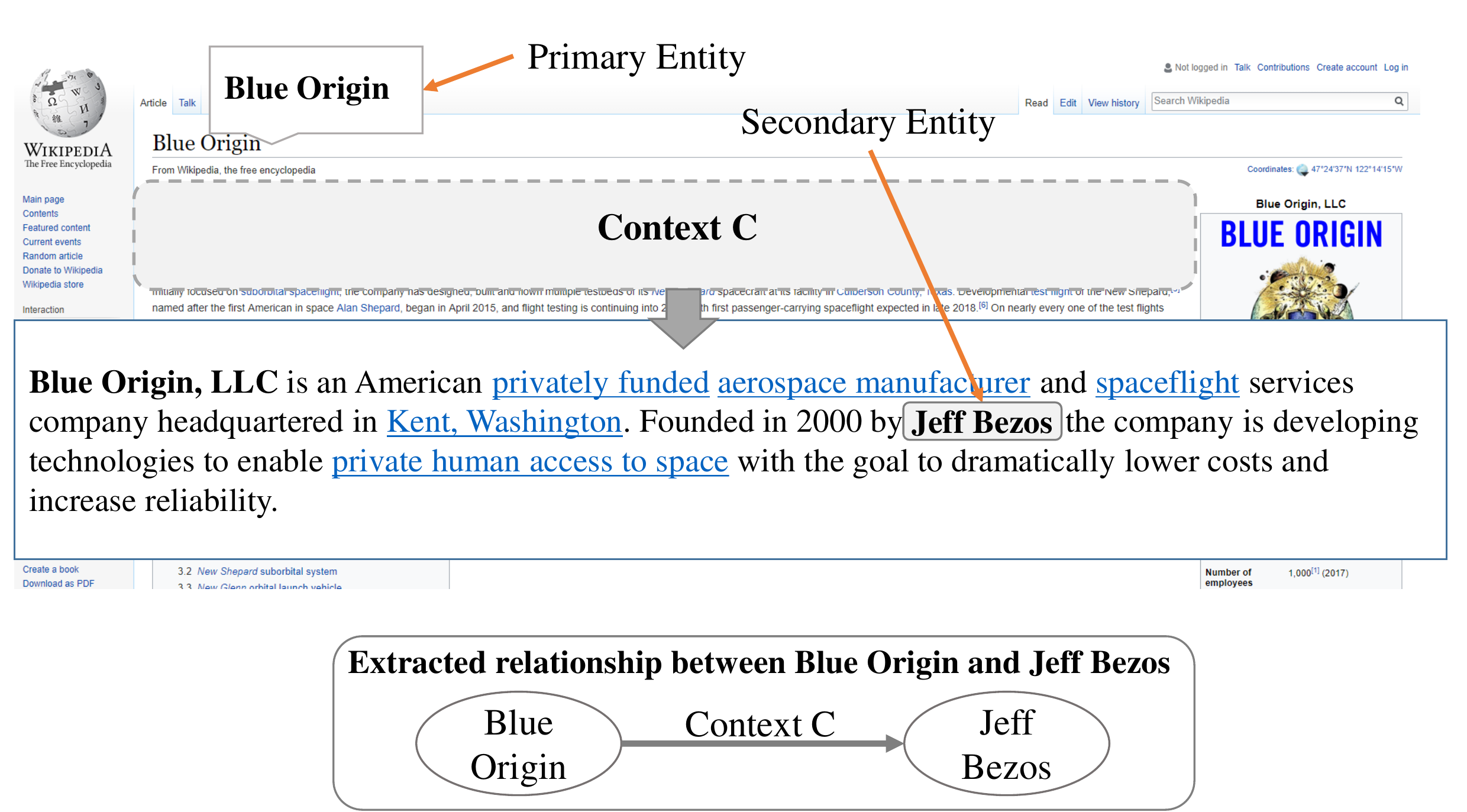}
    \caption{Example context triple extraction using `Blue Origin' Wikipedia entity page. In our experiments, for Wikipedia, entities are identified using hyperlinks and for Amazon reviews, aspect mining determines the entities. For other similar sources, a general purpose entity spotter can be used.}
    \label{fig_extraction}
\end{figure}

We extract entity context triples from textual knowledge bases that describe entities. These can be encyclopedic sources (e.g., Wikipedia) where each page describes a real world entity in detail or a data source where we can designate textual descriptions for relationships between entities (e.g., online review forums and product specifications). Online reviews and product specifications are different to a knowledge base like Wikipedia in the sense they do not have the level of structured information that Wikipedia has. But they are similar to a knowledge base in the sense each review or product specification page describes a certain product. Therefore, our approach is not confined to work with well-defined knowledge bases like Wikipedia only. The contrasting difference of context triples is that the relationship in a triple consists of a text description compared to a short label (see Figure~\ref{fig_ecg_triple}). Because of this special representation format, context triple extraction is easy and straightforward compared to traditional KG triple extraction as we do not need to align to a predefined set of relationships, which is one of the hard challenges of formal KG triple extraction. Specifically, this new extraction technique has certain advantages: (i) the relationship between a pair of entities can capture a rich and complex context via a text description, (ii) no need to align to a predefined set of relationships, and (iii) hence, the realization of the relationships can be application/domain specific (that is, the meaning can be inferred based on the application of the triple using for example attention).

The context triple extraction model is based on how a document describes an entity. In our approach, a document in an entity centric knowledge base (e.g., Wikipedia) describes an entity in detail using text. Different parts of the document present different facts of the entity, that can be the relations for the entity. We use this observation to extract context triples. See Figure~\ref{fig_extraction} for example for extracting an example context as the relationship between `Blue Origin' and `Jeff Bezos' entities using the Wikipedia page dedicated to `Blue Origin'. Let's consider a set of entities $E$ in a knowledge base $B$, where $P$ is the set of resource topic documents (topics/pages) that describe the entities $E$. For an entity $e_{p} \in E$, the resource document $p \in P$ that describes $e_{p}$ is called the topic. Let $E_{p}$ be the set of entities mentioned in the topic $p$. Then $e_{p}$ is called the \textit{primary entity} for the topic and entities in the set $E_{p} \setminus e_{p}$ are called \textit{secondary entities} for the topic $p$. We create directed edges from $e_{p}$ to each $e' \in E_{p} \setminus e_{p}$ with text surrounding the entity mention $e'$ up to a pre-determined token length $m$ as the relationship. A collection of such triples make up the ECG as outlined in Algorithm~\ref{alg_ecg}.

By following Algorithm~\ref{alg_ecg}, we create ECGs from semi-structured knowledge sources. We refer Wikipedia, online reviews, catalogues as semi-structured (do not have knowledge represented in structured format like triples but follow templates to organize text) whereas KGs as structured. Our approach requires identification of entities in text, which can be done with or without processing a KG (e.g., dictionary based, Wikipedia-like source based, other available mappings - refer to our experiments, or general purpose entity spotters like DBpedia Spotlight~\cite{mendes2011dbpedia} and Tagme~\cite{ferragina2010tagme}). That is, an entity can exist in a knowledge base like Wikipedia and be detected (a dedicated page per entity) even though Wikipedia is not a KG.

    \begin{algorithm}[H]
        \small
        \caption{Building Entity Context Graph}
        \label{alg_ecg}  
        \begin{algorithmic}[1]
        \STATE input: Collection $topic\_pages$
        \STATE output: Entity Context Graph $ecg$
        \FORALL{$p$ $\in$ $topic\_pages$}
        	\STATE List $entities$ $\gets$ extractEntityMentions($p$)
        	\FORALL{$e$ $\in$ $entities$}
        		\STATE Text $contex$t $\gets$ extractContext($p, e$)
                \STATE Node $primary\_entity$ $\gets$ getNode($p$)
                \STATE Node $secondary\_entity$ $\gets$ getNode($e$)
                \STATE Triple $c\_triple$ $\gets$ getTriple($primary\_entity$ \\ $, context,$ $secondary\_entity$)
                \STATE addToGraph($ecg, c\_triple$)
        	\ENDFOR
        \ENDFOR
        \STATE return $ecg$
        \end{algorithmic}
    \end{algorithm}

\subsection{Embedding Learning from ECGs}
The rich and explicit representation of structured data in KGs is used by many LR algorithms to compute entity embeddings. Since, we now have a graph representation of entities through ECG, we can compute embeddings by processing the ECG. But adapting existing LR techniques is not straightforward with ECG. This is because, relationships in context triples are not selected from a small fixed set of relationship labels and they consist of long text descriptions. Hence, they are often lexically unique to each other but may be semantically similar. For example, the presidency relation expressed in text for `Barack Obama' is different to `Donald Trump' in Wikipedia, but both texts link to `United States' entity and should be considered similar. Therefore, instead of initializing relationships to a fixed set of vectors, we encode them and learn this encoding step while training for embedding computation.

To generate the embeddings, we adapt TransE~\cite{bordes2013translating} algorithm, which is a simple translational model for triples. Further, there is no issue in substituting embedding lookup for relationships with the encoder. TransE minimizes the energy for a given triple (h,r,t) by enforcing the tail embedding $\boldsymbol{t}$ to be the closest to $\boldsymbol{h} + \boldsymbol{r}$ (i.e., $\boldsymbol{h} + \boldsymbol{r}$ $\approx$ $\boldsymbol{t}$). We adapted the original cost function of TransE by normalizing all the vectors for a triple and adding regularization, since entity and relationship vectors come from two different ways: initialization and encoding. The adapted cost function for a collection of triples $S$ (and corrupted triples $S'$) is shown in Equation~\ref{eq_1}. We follow the same head or tail replacement to generate the corrupted triples as proposed in TransE. Dissimilarity measure $d$ is set to either $L1$ or $L2$ distance.

\small
\begin{equation}
\small
\label{eq_1}
\mathcal{L'}=\sum\nolimits_{(h,r,t) \in S, (h',r,t') \in S'}^n
\frac{max(\gamma + d(\hat{\boldsymbol{h}} + \hat{\boldsymbol{r}}, \hat{\boldsymbol{t}}) - d(\hat{\boldsymbol{h'}} + \hat{\boldsymbol{r}}, \hat{\boldsymbol{t'}}), 0) \: + \: \mu}{n}
\end{equation}
\normalsize
where, $\gamma >$ 0 is the margin hyperparameter, $\mu$ is the regularization and $\hat{\boldsymbol{x}}$ is the normalized vector of $\boldsymbol{x}$. The adapted learning process is outlined in Algorithm~\ref{alg_embedding}.

\begin{algorithm}[H]
    \small
    \caption{Adapted Learning Algorithm}
    \label{alg_embedding}  
    \begin{algorithmic}[1]
    \STATE input: Training set $S$=\{$h,r,t$\}, entity and relationship sets $E$ and $R$, margin $\gamma$, embedding dimension $k$
    \STATE initialize weights of the encoder for relationship transformation
    \STATE e $\gets$ uniform(-$\frac{6}{\sqrt{k}}, \frac{6}{\sqrt{k}}$) for \: each \: e $\in$ E
    \STATE e $\gets$ e / $\|$e$\|$ for \: each \: e $\in$ E
    \LOOP
    \STATE $S_{batch}$ $\gets$ sample($S$, $b$) // $b$ is the batch size
    \STATE $T_{batch}$ $\gets$ $\emptyset$
        \FORALL{($h,r,t$) $\in$ $S_{batch}$}
            \STATE $\boldsymbol{r}$ $\gets$ encoder($r$)
            \STATE ($h'$,$r$,$t'$) $\gets$ sample($S'_{(h,r,t)}$)
            \STATE $T_{batch}$ $\gets$ $T_{batch}$ $\cup$ \{(($h$,$r$,$t$), ($h'$,$r$,$t'$))\}
        \ENDFOR
        \STATE $\mathcal{L'} = \sum_{((h,r,t),(h',r,t')) \in T_{batch}}\newline \frac{max(\gamma + f(\hat{\boldsymbol{h}} + \hat{\boldsymbol{r}}, \hat{\boldsymbol{t}}) - f(\hat{\boldsymbol{h'}} + \hat{\boldsymbol{r}}, \hat{\boldsymbol{t'}}), 0) \: + \: \mu}{|T_{batch}|}$ 
    \ENDLOOP
    \end{algorithmic}
\end{algorithm}

\subsubsection{Text Encoder}

We use a multi-layer Convolutional Neural Network (CNN) model to encode the relationship texts of context triples. The order of tokens in these relationships does not matter but their composition is important to decide the latent relationship label. We adapt a text encoding network to support this intuition.

Let $x_{i}$ $\in$ $\mathbb{R}^{d}$ be the $d$ dimensional word vector representation corresponding to the $i$-th word in the relationship text input of length $m$ (padded when necessary) where $x_{1:m}$ = $x_{1} \oplus x_{2} \oplus ... \oplus x_{m}$ and $\oplus$ is the concatenation operator. A convolution operation performs as a filter $w \in \mathbb{R}^{yd}$, which applies to a window size of $y$ words to produce a new feature. For each of these features, a non-linear activation $f$ is applied.

\begin{equation}
\label{eq_2}
    h_{i} = f(w \cdot x_{i:i+y-1} + b),
\end{equation}
where, $b \in \mathbb{R}$ is the bias. After applying convolution filters we get a feature map $h = [h_{1}, h_{2}, ..., h_{m-y+1}]$, where $h \in \mathbb{R}^{m-y+1}$. With padding, we can get the input length ($m$) as the dimension for $h$. Max pooling over window size $z$ on feature map $h$ is performed and outputs the feature map $\hat{h} = [\hat{h}_{1}, ..., \hat{h}_{m-y-z+2}]$. As we are interested in detecting phrases (that probably appear in different places in the text), we perform three convolution operations in parallel with different window sizes ($y=a$ $<$ $b$ $<$ $c$) to extract different word sequence features. Then we align the three feature maps by max pooling using a constant $\beta$ $>$ $c$. That is, we set $z$, the max pooling window size to be $\beta - a + 1$, $\beta - b + 1$, and $\beta - c + 1$ for the feature maps mined from $a, b,$ and $c$, respectively, to align them correctly and stack them. Then two additional convolution and max pooling layers are applied before connecting to a fully connected layer. These convolutions are important to combine the aligned features to glean insights from the relation text and the fully connected layer makes connections with distantly located features. Regularization is added to avoid overfitting.

\section{Evaluation}
\label{evaluation}

We evaluate and analyze our approach on different directions: (i) usefulness of ECG structure in learning embeddings, (ii) quality of the embeddings in comparison to the KG-based and adapted contextual language model-based embeddings, (iii) how context triples can complement KG triples to generate better (joint) embeddings, and (iv) analyze the feasibility and flexibility of our approach in a domain specific use case where finding a KG is impossible (and show that building one is unnecessary).

\subsection{Datasets and Implementation Details}

\begin{table}[]
\setlength\tabcolsep{2pt}
\footnotesize
\begin{tabular}{p{1.6cm}|p{0.3cm}p{0.3cm}p{0.3cm}p{0.3cm}p{0.75cm}p{0.35cm}}
\hline
\multicolumn{1}{l|}{Dataset} &    FB15k                       &     WN18     &                     Cities                            & Movies                        & Albums                            & Amazon \\ \hline
\# Entities                  & \multicolumn{1}{r}{11,757}  &       \multicolumn{1}{r}{40,943}      & \multicolumn{1}{r}{7,61,784}      & \multicolumn{1}{r}{69,875}    & \multicolumn{1}{r}{24,971}        & \multicolumn{1}{r}{37,156}      \\
\# Relationships            & \multicolumn{1}{r}{34,496}  &        \multicolumn{1}{r}{29,315}        & \multicolumn{1}{r}{8,94,884}      & \multicolumn{1}{r}{65,895}    & \multicolumn{1}{r}{18,487}        & \multicolumn{1}{r}{63,370}      \\
\# Train Triples         & \multicolumn{1}{r}{81,279}     & \multicolumn{1}{r}{1,05,665}             & \multicolumn{1}{r}{13,24,981}     & \multicolumn{1}{r}{1,65,714}  & \multicolumn{1}{r}{59,220}        & \multicolumn{1}{r}{2,09,383}      \\
\# Valid Triples       & \multicolumn{1}{r}{11,935}    & \multicolumn{1}{r}{N/A}             & \multicolumn{1}{r}{N/A}           & \multicolumn{1}{r}{N/A}       & \multicolumn{1}{r}{N/A}           & \multicolumn{1}{r}{27,917}       \\
\# Test Triples             & \multicolumn{1}{r}{13,465}  & \multicolumn{1}{r}{N/A}     & \multicolumn{1}{r}{N/A}           & \multicolumn{1}{r}{N/A}       & \multicolumn{1}{r}{N/A}           & \multicolumn{1}{r}{41,878}     \\ \hline
\end{tabular}
\caption{Statistics of the ECG graphs generated for each dataset. N/A means that we did not split the total number of triples and used all for training. The datasets have original entity sets where they were originally created and to get relationships for the entities to create ECGs, we use Wikipedia for FB15k, Cities, Movies, and Albums; WorNet for WN18; and product reviews for Amazon.}
\label{table_1}
\end{table}

We used three text-based knowledge sources, namely Wikipedia (English), WordNet (using word definition text), and Amazon reviews dataset~\cite{he2016ups}~\footnote{\url{https://cseweb.ucsd.edu/~jmcauley/datasets.html\#amazon_reviews}} to build the ECGs. The Amazon reviews dataset is quite different to Wikipedia, which is an encyclopedic textual knowledge base. It contains product reviews from users and we show how it could be processed to build an ECG to learn embeddings using the proposed approach. This also supports the fact that ECGs can be successfully built from different kinds of sources, not just encyclopedic and well known ones. The statistics of the generated ECGs are shown in Table~\ref{table_1}. From the listed datasets, FB15k is a sample of Freebase and WN18 is a sample of WordNet that were used in~\cite{bordes2013translating}, Cities, Movies, and Albums datsets were taken from a benchmark collection of datasets used for data mining and learning~\footnote{\url{http://data.dws.informatik.uni-mannheim.de/rmlod/LOD_ML_Datasets/data/}}, and Amazon is a sample extracted from Amazon reviews dataset. Note that we used the original entity sets described in FB15k, WN18, Cities, Movies, Albums datasets to create ECGs to compare against the baselines.

We set parameters for our evaluation as follows. We set the relationship text length $m$ to be 400, 120, and 100 for Wikipedia, Amazon reviews, and WordNet, respectively. For textual relationship extraction, first, we segmented paragraphs and then processed them to remove stop words. For each paragraph in Wikipedia, if the paragraph word length is larger than 400, we split the paragraph text (split into separate relationships) to have maximum of 400 words (120/100 for Amazon reviews/WordNet). If the extracted Wikipedia text (or review/definition text) is shorter than 400 (or 120/100 for Amazon reviews/WordNet), we zero pad the word entries to input into the encoder. For the CNN based encoder, we experimented on the learning rate for 0.1, 0.01, and 0.001; for margin $\gamma$, we experimented with values 1, 3, and 5; and used ReLU as the activation function ($f$). We used pre-trained word embedding model on Wikipedia and set the dimension ($d$) to be 100 and did not update them in RL process. For the first layer in the CNN (using 1d CNN), we set $a,b$ and $c$ to be 3, 5 and 7, respectively, and the number of filters 128, 64, and 32 in the order. We set $\beta$=10, and stride length 1 for convolutions and max pooling windows. For the second convolution layer, we set window size 5, stride 2, number of filters 280 and for max pooling, we set window size 3 and stride 1. For the third layer convolution, we set window size 5, stride 3, number of filters 350 and for max pooling, we set window size 3 and stride 1. These values were selected using validate sets. All models trained for 1000 iterations. Code will be made publicly available in the final version.


\subsection{Evaluation I}
We compare our approach's embeddings to RDF2Vec~\cite{ristoski2016rdf2vec} and adapted ERNIE~\cite{zhang2019ernie} based embeddings. Triples for RDF2Vec were taken from DBpedia and Wikidata KGs whereas our approach and ERNIE-based system used Wikipedia as the source. All three systems used the same entity sets.
\subsubsection{ECG and Learning} First, it is important to realize how well the extracted context triples can be used to model entities to capture their semantics, similar to a KG. For this, we use the link prediction task~\cite{bordes2013translating} on the FB15K dataset. Link prediction is performed by either holding out head or tail entities from the test triples and testing the held out entity ranking in the corpus. We achieved 70\% accuracy on Hits@10 and mean rank of 386 for predicting heads and tails on the test set using $k$=200 and $\gamma$=5. This accuracy on Hits@10 is comparatively high~\footnote{See TransE that reported 47.1\% accuracy in~\cite{bordes2013translating} on Freebase KG.} and suggests that the proposed approach can represent entities with their latent relationships using text descriptions and also learn their embeddings. We also empirically tested that the sequence of information is not that important in learning the embeddings. We encoded the relation text using attention based LSTM model (similar to ~\cite{yang2016hierarchical}) and it achieved 65\% for Hits@10. 

\subsubsection{Quality Comparison} We compare ECG embeddings to RDF2Vec (KG-based approach) and adapted ERNIE (contextual language model-based) embeddings. The embedding qualities are compared based on a set of classification tasks defined on the set of datasets proposed by the RDF2Vec~\cite{ristoski2016rdf2vec}. They considered higher classification accuracy to reflect the quality of the generated embeddings. We consider RDF2Vec as the KG-based baseline (also because of its evaluation on these external datasets) and ERNIE which is a pure language-based approach as the text-based baseline. We do not consider K-BERT~\cite{liu2020k} as a language model baseline as it needs KG triples. ERNIE is built on top of BERT~\cite{devlin2018bert} and uses KG entities together with words to improve the contextual language modeling. ERNIE was not designed to learn entity embeddings and hence, it loads pre-trained KG embeddings and does not update them in the training phase. We adapted ERNIE by pre-training using entire English Wikipedia and enabled updating entity embeddings after initializing them by picking a uniform distribution in the range (-1,+1). We used batch size of 24, learning rate of 1e-4, and performed over 500k update steps~\footnote{https://github.com/thunlp/ERNIE}. The RDF2Vec embeddings are generated by processing DBpedia and Wikidata, which are rich and encyclopedic KGs. Cities dataset contains quality of living for some cities, and Movies and Albums represent average ratings for all time reviews. The classification tasks categorize the entities based on their associated values labelled as `low', `medium', and `high'. We generated three ECGs using Wikipedia for these entities (see Table~\ref{table_1}). ERNIE did not fit into a 32GB gpu for $k$=500 and hence, we report only $k$=200.

\begin{table*}[]
\small
\begin{tabular}{p{5cm}|p{0.8cm}p{0.8cm}p{0.8cm}|p{0.8cm}p{0.8cm}p{0.8cm}|p{0.8cm}p{0.8cm}p{0.8cm}}
\hline
Strategy / Dataset & \multicolumn{3}{c|}{Cities}   & \multicolumn{3}{c|}{Movies} & \multicolumn{3}{c}{Albums}   \\
                        & NB        & KNN       & SVM       & NB        & KNN       & SVM       & NB       & KNN     & SVM      \\ \hline
DBpedia-RDF2Vec-CBOW k=200 l=4   & 59.32     & 68.84     & 77.39     & 65.60     & 79.74     & 82.90     & 70.72    & 71.86   & 76.36    \\
DBpedia-RDF2Vec-SG k=200 l=4     & 60.34     & 71.82     & 76.34     & 65.25     & 80.44     & 83.25     & 68.95    & 73.89   & 76.11    \\
DBpedia-RDF2Vec-CBOW k=500 l=4   & 59.32     & 71.34     & 76.37     & 65.65     & 79.49     & 82.75     & 69.71    & 71.93   & 75.41    \\
DBpedia-RDF2Vec-SG k=500 l=4     & 58.34     & 72.84     & 76.87     & 65.45     & 80.14     & \textbf{83.65}     & 70.41    & 74.34   & \textbf{78.44}    \\ \hline
Wikidata-RDF2Vec-CBOW k=200 l=4   & 68.76     & 57.71     & 75.56     & 51.49     & 52.20     & 51.64     & 50.86    & 50.29   & 51.44    \\
Wikidata-RDF2Vec-SG k=200 l=4     & 72.58     & 57.53     & 75.48     & 69.53     & 70.14     & 75.39     & 60.32    & 62.03   & 64.76    \\
Wikidata-RDF2Vec-CBOW k=500 l=4   & 68.24     & 57.75     & \textbf{85.56}     & 49.22     & 48.56     & 51.04     & 53.08    & 50.03   & 52.33    \\
Wikidata-RDF2Vec-SG k=500 l=4     & \textbf{83.20}     & 60.72     & 79.87     & 71.10     & 70.19     & 76.30     & 55.31    & 58.92   & 63.42    \\ \hline \hline
Our Approach k=200 $\gamma$=3   & 71.80     & 73.43     & 79.01     & \textbf{81.08}     & 80.12     & 81.38     & 68.00    & 74.48   & 74.70    \\
Our Approach k=200 $\gamma$=5   & 69.59     & \textbf{77.14}     & 79.88     & 80.52     & \textbf{80.72}     & 80.98     & 71.07    & 74.71   & 72.72    \\
Our Approach k=500 $\gamma$=3   & 77.08     & 65.89     & 75.84     & 80.78     & 79.20     & 80.72     & 70.62    & 74.59   & 76.49    \\
Our Approach k=500 $\gamma$=5   & 72.34     & 61.21     & 77.66     & 80.27     & 80.42     & 81.18     & \textbf{71.90}    & \textbf{75.09}   & 77.20 \\ \hline
ERNIE based (BERT based) k=200 *   & 47.02     & 55.57     & 61.23     & 55.03     & 50.94     & 52.88     & 49.90    & 53.60   & 52.34 \\ \hline
\end{tabular}
\caption{Accuracy \% of the classification tasks. NB, KNN, and SVM, refers to Naive Bayes, K Nearest Neighbor, and Support Vector Machines. k, l, and $\gamma$ refer to embedding dimension, path length, and margin. * could not fit ERNIE k=500 into 32GB GPU. The expectation is to get closer to KG-based results.}
\label{table_3}
\end{table*}

The results are shown in Table~\ref{table_3} for the classification tasks on the datasets by using Naive Bayes (NB), K Nearest Neighbor (KNN), and Support Vector Machines (SVM) as the classifiers. For SVM, we optimized the parameter $C$ in the range \{1, 10, 10$^{3}$, 10$^{5}$, 10$^{7}$\} and set gamma to 0.01. The results shown are the \% of the correctly predicted labels and higher values mean the learned embedding vectors are of good quality (refer ~\cite{ristoski2016rdf2vec}). We performed 10-fold cross validation for all the datasets. Embeddings generated by our proposed approach shows \textit{comparable} and \textit{high quality} results on the classification tasks against the RDF2Vec baseline that purely used DBpedia and Wikidata KGs, whereas our approach generated the embeddings using the automatically extracted ECGs from Wikipedia. Note that the goal is to achieve results closer to a KG-based baseline using a text only approach. ERNIE-based embedding results are lower than ours and the KG-based RDF2Vec. This may be because even though transformers can successfully learn contextual representation of words, it may be difficult to learn representations for entities using the same process. This also demonstrates why relationships are important in capturing meaning of entities~\cite{sheth2004relationships}. Comparing to these selected baselines, our approach works better in utilizing text only data to generate the embeddings.

\subsection{Evaluation II}

\begin{table}[]
\footnotesize
\begin{tabular}{p{1.1cm}|p{1.9cm}|l|l|l|l}
\hline
\multicolumn{2}{l|}{\multirow{2}{*}{Models}} & \multicolumn{2}{l|}{FB15K} & \multicolumn{2}{l}{WN18} \\ \cline{3-6} 
\multicolumn{2}{l|}{}                        & MR          & Hits@10        & MR        & Hits@10     \\ \hline
\multirow{5}{*}{TransE}        & TransE       & 125         & 47.1         &  251      & 89.2      \\ 
                               & TransE-TEKE      & 79          & 67.6         &  127      & 93.8      \\ 
                               & TransE-ATE       & 89          & 57.1         &  158      & 91.7      \\ 
                               & TransE-AATE      & 76          & 76.1         &  123      & 94.1      \\ 
                               & KG+ECG         & 48          & 80.0         &  688      & 92.5      \\ \hline
\multirow{5}{*}{TransH}        & TransH       & 84          & 58.5         &  303      & 86.7      \\ 
                               & TransH-TEKE      & 75          & 70.4         &  128      & 93.6      \\ 
                               & TransH-ATE       & 80          & 68.2         &  167      & 92.5      \\ 
                               & TransH-AATE      & 73          & 74.6         &  132      & 94.0      \\ 
                               & KG+ECG         & 62          & 82.2         &  754      & 92.7      \\ \hline
\multirow{5}{*}{TransR}        & TransR       & 78          & 65.5         &  219      & 91.7      \\ 
                               & TransR-TEKE      & 79          & 68.5         &  203      & 92.3      \\ 
                               & TransR-ATE       & 80          & 67.2         &  210      & 92.1      \\ 
                               & TransR-AATE      & 77          & 69.4         &  185      & 93.7      \\ 
                               & KG+ECG         & 100         & 69.5         &  552      & 93.2      \\ \hline
\multirow{4}{*}{ComplEx}       & ComplEx      & 78          & 84.0         &  219      & 94.7      \\ 
                               & ComplEx-ATE       & 61          & 86.2         &  217      & 94.7      \\ 
                               & ComplEx-AATE      & 52          & 88.0         &  179      & 94.9      \\ 
                               & KG+ECG         & 62          & 86.7         &  784      & 94.8      \\ \hline
\end{tabular}
\caption{Link prediction \% results (text-based baseline TEKE~\cite{wang2016text} and ATE~\cite{an2018accurate} and AATE~\cite{an2018accurate} taken from~\cite{an2018accurate}). MR - Mean Rank. KG+ECG is our approach. ECG uses TransE optimization only in all models.}
\label{table6}
\end{table}

In this experiment, we evaluate the versatility of ECGs by using them to complement KGs in KG+text-based embedding learning. KGs are often incomplete and costly to update continuously. But our method captures context triples from text sources in relatively easy and automated manner and hence, it can easily capture evolving or complementary information to a KG. State-of-the-art KG embedding learning approaches use text to complement KG information by aligning supplementary text data to KGs~\cite{wang2016text,an2018accurate}. In contrast, we do not do any alignment. 

We adapt our approach to process KG and ECG graphs by learning embeddings with two networks: (i) one for KG triples using existing approaches, and (ii) one for ECG adapting TransE with the dynamic encoding of textual relationships. Both networks share the same entity embedding space but optimize on their own optimization functions. We run the two networks one after the other in each iteration in the training phase. Once the training is complete, we only use the KG network to predict KG triples and the results are shown in Table~\ref{table6}. We compare our results against two KG+text baselines TEKE~\cite{wang2016text} and ATE and AATE~\cite{an2018accurate}. We generated the ECG for WN18 dataset by putting the WordNet word definition of the head of each WN18 triple as the relation in context triples. ECG for FB15k is as described in the earlier experiments. Then for our KG+ECG setting, we use original KG dataset for KG network and the generated ECG (statistics in Table~\ref{table_1}) for the other network. The KG+ECG joint learning improved results on FB15k dataset. The improvements are more apparent for TransE and TransH baselines. This is because, the baseline systems have used the respective optimization functions to update weights (e.g., TransE, TransR, TransH, and ComplEx) whereas we consistently used only TransE optimization to update the ECG network in all joint models. Adapting the same advanced optimization function used in the KG network to the ECG network in the joint model can improve our results. For WordNet, all the systems including KG only methods achieved very high accuracy numbers mainly due to the inherently simple structure (and limited relationships) in the dataset. Moreover, word definitions in WN18 which are used to generate the ECG are very short (e.g., often around 10 words) and hence, adding very little additional context information to learn from the generated ECG. Our approach requires rich context information to learn better embeddings. Also note that our approach achieves higher accuracy without any alignment of text to the KG triples and did not use any attention mechanism. Our approach outperformed almost all systems (all in FB15K) when considering no extra attention mechanism and incorporating such mechanism (as in AATE) may lead to further improvements.

\subsection{Use Case Analysis - Domain Specific ECG Building and Learning}
We emphasis the flexibility and usefulness of our approach with a domain specific use case, where we avoid the need to process a KG. We show how to use our method to compute embeddings for product `aspects' in order to measure their similarity. Aspects are semantically coherent attributes related to a product that reviewers may hold an opinion about~\cite{zhang2014aspect}. Aspects have very specific relationships to types of products. For example, electronic products can have a `battery life aspect', while clothing can have a `comfortable fit' aspect. Understanding the relationships between aspects, products, and other aspects is important because, it can provide more depth for real-world NLP applications in recommendations, advertisements, and question answering systems. As we have discussed earlier, building KGs are time consuming and labor intensive, and we could not find a comprehensive KG for product aspects like DBpedia or Freebase (mainly because this is application specific). Therefore, it is impossible to use an existing KG based learning approach to generate embeddings for the aspects. We extract context triples from product reviews in the form `user'-`review text'-`aspect' to construct an ECG (where users and aspects are entities). For a review, a context triple can be generated by linking the user and a mined aspect from the review and having the review text as the relationship.

\subsubsection{Dataset Creation and Aspect Mining} We mined the top 10 frequent aspects for the product reviews on a subset of 281 product categories from the top 2 levels of the Amazon category hierarchy. We selected about half of the categories at that depth as domains of interest using keywords: Books, Cell Phones, Music, Movies \& TV, Sports, Games, Clothing, and Home \& Kitchen. We randomly selected reviews so that each aspect would have 50 reviews, if possible, in making the dataset for aspects. We treated each aspect and user as entities in the ECG. The aspects were mined using the Clustering for simultaneous Aspect and Feature Extraction (CAFE)~\cite{chen2016clustering} algorithm. An \textit{unsupervised} approach like CAFE or topic modeling is appropriate for this dataset as the number of categories is large and each product type has a wide variety of different aspects.

\begin{table}[]
\centering
\small
\begin{tabular}{l|l|l}
\hline
\multicolumn{1}{c|}{Settings}     & \multicolumn{1}{|c|}{Mean Rank}         & \multicolumn{1}{c}{Hits@10} \\ \hline
$\gamma$=3, k=200            & 293                     & 62                 \\ \hline
$\gamma$=5, k=200            & 262                     & 67                 \\ \hline
\end{tabular}
\caption{Link prediction \% on our ECG generated from Amazon product review dataset.}
\label{table_4}
\end{table}

\begin{table*}[]
\begin{tabular}{p{6.6cm}|p{0.8cm}|p{6.6cm}|p{0.7cm}}
\hline
\multicolumn{1}{c|}{Aspect 1}     & \multicolumn{1}{|c|}{Category}                               & \multicolumn{1}{c|}{Aspect 2} & \multicolumn{1}{|c}{Category}                       \\ \hline
Information: \{Information, fact, truth, advice, content, message, info, answers \} & Books & School : \{ school, class, college, students, lesson, teacher, teach, moral, learns\} & eBooks \\ \hline
Clothes: \{ shirt, pants, hat, socks, jacket, shorts, shoes, underwear, gloves, cap, boots, slacks, scarf, ...\} & Clothing & Body Part : \{ hand, hands, feet, legs, wrist, arms, waist, shoulder, leg, finger, fingers, arm, neck, thumb, butt\} & Sports \\ \hline
Season : \{ time, spring, summer, winter, ice, season\} & Sports                          & Fabric : \{fabric, cotton, denim, wool, cloth, silk, polyester, ... \} & Clothing\\ \hline
Fast/Slow Pace : \{pace, step, slow, fast\} &  Books                            & Fun : \{enjoy, exciting, engaging, interesting, fun ...\} & Movies                                                    \\ \hline
\end{tabular}
\caption{A few related aspect pairs computed using cosine similarity of the ECG based aspect embedding vectors. Words in brackets are the ones that define the respective aspects. ECG is generated using Amazon reviews and CAFE algorithm is used to identify aspects. Context triples are in the form `user-review-aspect'.}
\label{table_5}
\end{table*}

\subsubsection{Analysis} We performed the link prediction task for aspects on the ECG and the experiment results are shown in Table~\ref{table_4}. They suggest that our learning approach captured triple patterns successfully in the ECG (which is built by capturing user reviews for aspects and shows similar performance to the first part of the Experiment I) and the method can be used to model relationships in various types of datasets, not restricting to well-known ones like Wikipedia.

We discovered links between different aspects by measuring their similarity in the embedding space that we use in a recommendation application (we omit further details of the application as it is out of scope). Table~\ref{table_5} shows a few example links computed by the cosine similarity of the aspect embeddings. Identifying links between cross-domain product aspects is difficult using lexical similarity measures because the cross-domain product aspects share little lexical similarity as we mined them from application specific disjoint sub domains of the Amazon product category tree. But the ECG embedding similarity-based linking can find different kinds of relationships between aspects in different product categories. Take for example a basic analogy, the pace of the book is like the excitement level of a movie.  This is true, but we can find other relationships as well. For example, `Clothes' and `Body Part' aspects have different words (shown in brackets in Table~\ref{table_5}) that describe the aspects and they show no similarity on the lexical level but they can be linked because people wear those clothes on those body parts. Similarly, certain kinds of `Fabric' are better given the `Season'. In `School' you learn `Information'. 
We used these insights to connect related products and categories by analyzing embeddings of the aspects generated by our framework, which would have been quite difficult otherwise.

\section{General Discussion}
\label{general_discussion}
Our approach can generate quality entity embeddings by processing ECGs that can be easily generated automatically from textual knowledge sources. This is versatile because we can use ECGs where there exists no KG. Moreover, our context triples can model a complex relationship description using one triple, making the representation and extraction of complex relationships straightforward whereas, a formal triple representation may have to follow a complex ontology pattern to represent the same information (e.g., see Figure~\ref{fig_ecg_triple}). Extracting these complex relationships for KGs is nontrivial and hence, some of the popular KGs do not include them (e.g., DBpedia). Moreover, when there is a KG available, we can use the automatically generated ECGs to learn even better quality KG-based embeddings. On the other hand, context triple representation can be easily used to model naturally text heavy real-world applications like user conversations in dialogues and intelligent assistant systems. We intend to apply our approach in personal assistant conversation understanding in near future. Moreover, we can also try to extend the embedding learning approach to have a semantic layer (such as dependency parse results and tags in the order they appear) of the text as a separate input to improve on the embedding quality by allowing the network to better identify latent relationships in the text descriptions.

Our approach showed higher Hits@10 values (70\% for FB15k and 67\% for Amazon reviews), indicating it predicted more held out heads and tails of the test triples correctly. However, the mean ranks are also higher. This indicates that while it predicted more correct ones at the top of the list, few were at the bottom of the list. This is not surprising because, the learning happens on text-based and less formally defined relationship guided context triples. In fact, by refining the context text extraction using a sliding window rather than a paragraph can improve the quality of relationship representation, which can in turn improve the embeddings. On the other hand, the experiments showed that our method achieved better results over the adapted ERNIE approach, which is built on top of transformer-based BERT that learns from large text corpus. Both our approach and ERNIE processed the same text corpus Wikipedia and the better results in our approach is due to the graph representation of entities.

Our proposed approach works with any semi-structured entity centric source, meaning that entities are described in isolation as in for example Wikipedia and online reviews. It is possible to create graphs for sources like NEWS articles where each document may describe more than one entity (e.g., headline/title can contain more than one entity often). Creating context triples by linking all primary entities to secondary entities may further add noise to the already relaxed (compared to KG triple representation) ECG representation. In future, we plan to investigate in this direction by applying topic detection for the article (e.g., in NEWS article title) and then create the context triples. This way of processing widens the applicability of our approach.

When building a KG, most of the effort is focused on triple extraction. This involves entity and relationship detection and connecting the entities and relationships to form correct triples. In our proposed approach, we require entity detection but do not need to link relations or find correct triples conforming to a schema/ontology. This makes our ECG creation approach relatively simple and let the relation encoder dynamically encode the textual relationships. In our experiments, Wikipedia entities were identified by their hyperlinks and aspects were discovered and disambiguated by the aspect mining algorithm. Moreover, entity detection can be performed using a general purpose tool like DBpedia spotlight, making domain and application specific ECG creation easy when entities are not already labeled as found in Wikipedia. Further, since we follow a simple context extraction to represent relationships between entities, the actual relationship representation may be mixed with some other relationships (i.e., one-to-many). A better context extraction method for relationships (e.g., learning-based method to segment most relevant text for a secondary entity in a text description) can improve the approach and we plan to focus more on this direction in the near future.

\section{Conclusion}
\label{conclusion}
We proposed a novel entity embedding learning approach based on ECGs, which can be easily built from semi-structured data sources automatically. We conducted a comprehensive evaluation that spanned over several directions, used three different knowledge sources and four different KGs, and built six different ECGs. We showed that the: (i) adapted embedding approach works as well as a state-of-the-art RL approach on KGs, (ii) quality of our embeddings is comparable to a state-of-the-art KG-based embedding approach and better than a transformer-based language + entity model, (iii) ECGs can enhance joint KG+text embeddings, and (iv) ECGs can be generated from different data sources and defined to capture application specific needs. We believe that our approach can be useful in many challenging and different use cases and ECGs can be helpful in numerous applications such as search and retrieval, contextual ranking, and recommendation.

\newpage
\bibliographystyle{ACM-Reference-Format}
\bibliography{kge_www21}

\end{document}